\pdfoutput=1

\documentclass[11pt]{article}

\usepackage{ACL2023}

\usepackage{times}
\usepackage{latexsym}
\usepackage{graphicx}
\usepackage{booktabs}
\usepackage{multirow}
\usepackage{amsmath}
\usepackage{amsfonts}
\usepackage{subcaption}
\usepackage{tikz}
\usepackage{float}
\usepackage{tcolorbox}
\usepackage{float}
\usepackage{soul}
\usepackage{algorithm}
\usepackage[linguistics]{forest}
\usepackage{algpseudocode}
\usetikzlibrary{arrows.meta}
\usepackage{pgfplots}
\usetikzlibrary{bayesnet}

\usepackage[T1]{fontenc}

\usepackage[utf8]{inputenc}

\usepackage{microtype}

\usepackage{inconsolata}

\usepackage{caption}
\usepackage{titlesec}

\title{Quasi-symbolic Semantic Geometry over \\ Transformer-based Variational AutoEncoder}

\author{Yingji Zhang$^{1\dagger}$,~ Danilo S. Carvalho$^{1,3}$,~ Andr\'{e} Freitas$^{1,2,3}$ \\
  $^{1}$ Department of Computer Science, University of Manchester, United Kingdom\\
  $^{2}$ Idiap Research Institute, Switzerland\\
  $^{3}$ National Biomarker Centre, CRUK-MI, Univ. of Manchester, United Kingdom\\
  \texttt{\{firstname.lastname\}@[postgrad.]$^{\dagger}$manchester.ac.uk}}

\begin{document}
\maketitle

\begin{abstract}
Formal/symbolic semantics can provide canonical, rigid controllability and interpretability to sentence representations due to their \textit{localisation} or \textit{composition} property. How can we deliver such property to the current distributional sentence representations to better control and interpret the generation of language models (LMs)? In this work, we theoretically frame the sentence semantics as the composition of \textit{semantic role - word content} features and propose the formal semantic geometrical framework. To inject such geometry into Transformer-based LMs (i.e. GPT2), we deploy a supervised Transformer-based Variational AutoEncoder, where the sentence generation can be manipulated and explained over low-dimensional latent Gaussian space. In addition, we propose a new probing algorithm to guide the movement of sentence vectors over such geometry. Experimental results reveal that the formal semantic geometry can potentially deliver better control and interpretation to sentence generation.
\end{abstract}

\section{Introduction}

Language Models (LMs) have provided a flexible scaling-up foundation for addressing a diverse spectrum of tasks \cite{touvron2023llama}. Nonetheless, the question remains: can we develop language representations/models that offer more granular levels of control and interpretation from the perspective of ``formal/structural'' semantics? Addressing this question will enable us to enhance the controllability, interpretability, and safety of LMs.

Formal semantics, which provides a canonical, granular, and rigid representation, have been investigated for thousands of years with well established theoretical frameworks, such as Montague Semantics \cite{dowty2012introduction}, Davidsonian Semantics \cite{davidson1967logical}, Semantic Role Labelling (SRL, \citet{palmer2010semantic}), and Argument Structure Theory (AST, \citet{jackendoff1992semantic}).  One typical characteristic of such formal semantics is the \textit{localisation} or \textit{composition} property. For example, in the sentence: \textit{animals require oxygen for survival}, the words are functionally combined into sentence semantics: 
$\lambda x (\text{animals}(x) \rightarrow \text{require}(x, \text{oxygen}))$
where $x$ is the variable of any entity within a logical structure. In this case, we can localise the sentence semantics by replacing $x$ with \textit{birds}, etc. This localised process indicates the interpretation in Cognitive Science \cite{lees1957syntactic,smolensky2006harmony}. However, such localisation is precisely what current distributional semantics lack, thereby limiting their controllability and interpretability.
\begin{figure}[t]
    \centering
    \includegraphics[width=\columnwidth]{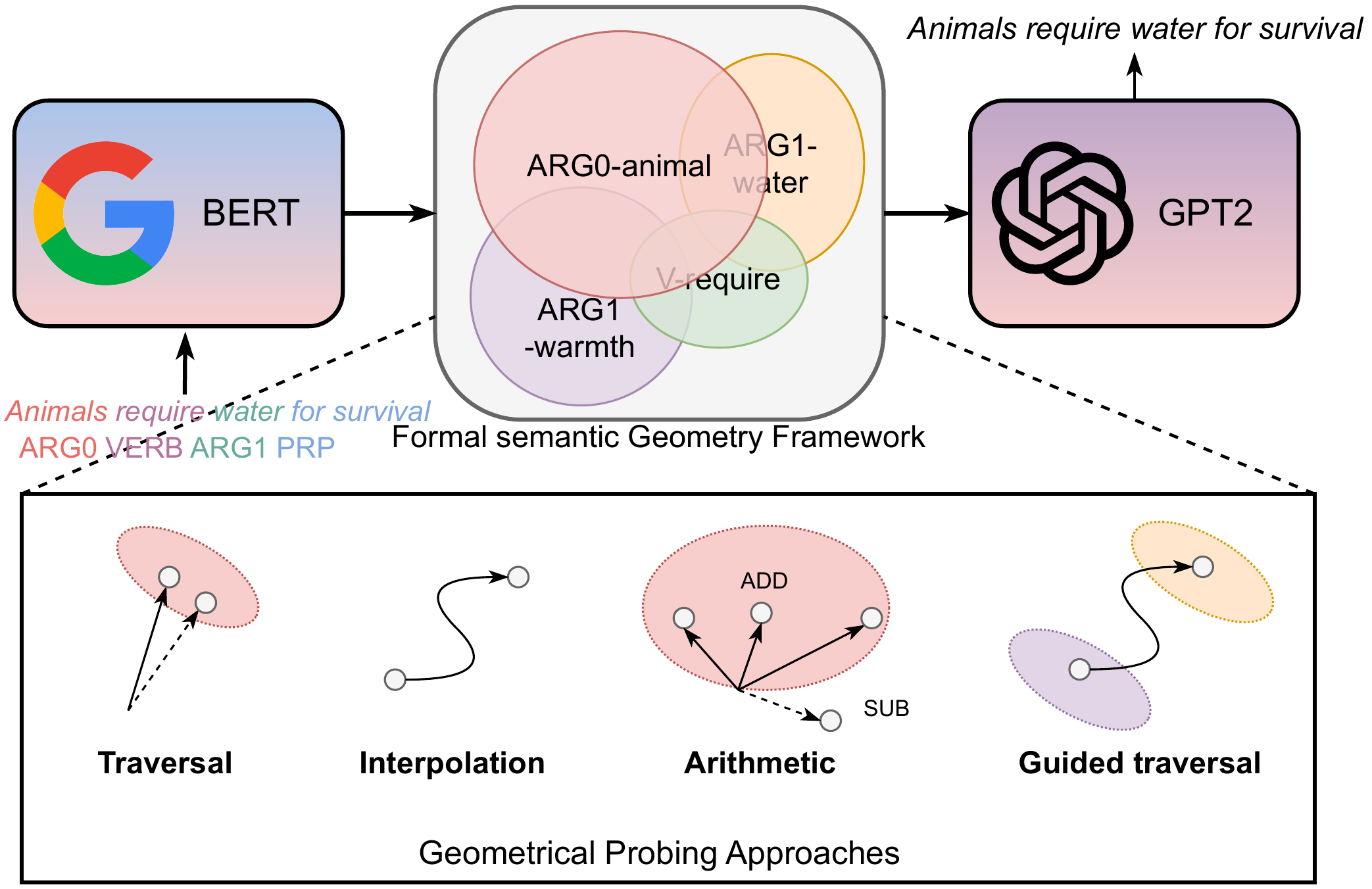}
    \caption{Overview: latent sentence semantics can be decomposed into \textit{semantic role- word content} features.}
    \label{fig:e2e_arch}
\end{figure}

Disentanglement \cite{bengio2013deep}, which refers to the feature-dimension alignment, can potentially provide such localisation, which has been widely investigated to localise image features, such as \textit{nose} in facial images  \cite{esser2020disentangling,jeon2019efficient, liu2021smoothing}. In Transformers \cite{vaswani2017attention}, however, token embeddings, residual stream, and attention have the \textit{polysemanticity} phenomenon \cite{elhage2022superposition}, meaning that multiple dimensions contribute to a feature. Although some prior studies explored the possibility of language disentanglement, most are focused on coarse-grained/task-specific semantic features, such as sentiment, within the context of style-transfer tasks \cite{john2019disentangled,bao2019generating,hu2021causal,vasilakes-etal-2022-learning,gu-etal-2022-distributional,liu-etal-2023-composable,gu-etal-2023-controllable}. 

 In this work, we focus on the localisation of \textit{general} semantic features of sentences over distributional space to shorten the gap between deep latent semantics and formal linguistic representations \cite{10.3115/1075218.1075283,banarescu2013abstract,mitchell2023we}, integrating the flexibility of distributional-neural models with the properties of linguistically grounded representations, facilitating both interpretability and generative control from the perspective of formal semantics. We specifically choose the conceptual dense explanatory sentences from WorldTree \cite{jansen2018worldtree} due to their clear formal semantic representation designed in the explanatory, cognitive reasoning task.

In the NLP domain, Variational AutoEncoders (VAEs, \citet{https://doi.org/10.48550/arxiv.1312.6114}) have been recognized as a prominent foundation for investigating generation control and interpretation through the observable low-dimensional smooth and regular latent spaces (e.g., std Gaussian space) \cite{john2019disentangled,li-etal-2022-variational-autoencoder,bao2019generating,mercatali2021disentangling, felhi2022towards,vasilakes-etal-2022-learning}. Therefore, we probe the localisation property of formal semantics over latent sentence spaces under VAE architecture. Specifically:

\textbf{(1)} We first propose a geometrical framework to present the formal semantic features of sentences as \textit{semantic role - word content} pairs (denoted as role-content) from the perspective of AST \cite{jackendoff1992semantic} within the compositional distributional model \cite{clark2008compositional}. Subsequently, \textbf{(2)} we introduce a supervised approach for learning the role-content features of explanatory sentences in latent spaces. \textbf{(3)} Additionally, we contribute to a method to control sentence generation by navigating the sentence vectors across different role-content features within our geometric framework. \textbf{(4)} Our findings reveal that the role-content features are encoded as a convex cone in the latent sentence space (Figure \ref{fig:e2e_arch}). This semantic geometry facilitates the localisation of sentence generation by enabling the manipulation of sentence vectors through traversal and arithmetic operations within the latent space. 
\section{Related work} \label{sec:related}

\paragraph{Formal-distributional semantics.} Integrating distributional semantics with formal / symbolic semantics is challenging due to the difficulty of optimisation over discrete space \cite{van2023nesi}. In the Reasoning domain, existing approaches usually perform symbolic behaviour via explicitly symbolic representation injection, including graph \cite{khashabi2018question,khot2017answering,jansen2017framing,thayaparan2021explainable}, linear programming \cite{valentino2022case,thayaparan2024differentiable}, adopting iterative methods, using sparse or dense encoding mechanisms \cite{valentino2020explainable,lin2020differentiable,valentino2022hybrid,bostrom-etal-2021-flexible}, or synthetic natural language expression \cite{clark2020transformers,yanaka-etal-2021-sygns,fu2024exploring}, among others. Comparatively, we explore the formal semantic property over distributional semantics via latent sentence geometry, which can potentially deliver better interpretation and control to current LMs.

\paragraph{Language geometry.} There is a line of work that studies the geometry of word and sentence representations \cite{arora-etal-2016-latent,mimno-thompson-2017-strange,ethayarajh-2019-contextual,reif2019visualizing,li2020sentence,chang2022geometry,jiang2024uncovering}. E.g., $king - man + woman = queen$, which the word vectors can be manipulated with geometric algebra. This phenomenon indicates the linear subspaces in language representations, similar features are encoded as a close direction in latent space, which has been widely explored ranging from word \cite{mikolov2013distributed} to sentences \cite{ushio2021bert}, Transformer-based LMs \cite{merullo2023language,hernandez2023linearity}, and multi-modal models \cite{trager2023linear,huh2024platonic}. Under the linear subspace hypotheses, a significant work explored the interpretability \cite{li2022emergent,geva2022transformer,nanda2023emergent} and controllability \cite{trager2023linear,merullo2023language,turner2023activation} of neural networks. In this work, we emphasise the formal semantic geometry for bridging the distributional and formal semantics, which is currently under-explored.

\paragraph{Language disentanglement.} Disentanglement, refers to separating features along dimensions \cite{bengio2013deep}, leading to clear geometric and linear representations. In the NLP domain, prior studies explored the disentanglement between specific linguistic perspectives, such as sentiment-content \cite{john2019disentangled}, semantic-syntax \cite{bao2019generating}, and negation-uncertainty \cite{vasilakes-etal-2022-learning}, or syntactic-level disentanglement \cite{mercatali2021disentangling, felhi2022towards}. However, those approaches focused on disentangling coarse-grained/task-specific semantic features. 
In this work, we contribute to a new lens on the disentanglement (separation) of ``general'' sentence features from the perspective of formal semantics.

\section{Formal Semantic Geometry} \label{sec:latent_props}
In this section, we first define the sentence semantic features as \textit{semantic role - word content} from the perspective of formal semantics. Then, we link the semantic features with distributional vector spaces in which each \textit{semantic role - word content} is encoded as a convex cone, as shown in Figure \ref{fig:e2e_arch}. 
\paragraph{Formal semantic features.} For formal / structural semantics, \textit{Argument Structure Theory (AST)} \cite{jackendoff1992semantic, levin1993english, rappaport2008english} provides a model for representing sentence structure and meaning of sentences in terms of the interface between the their syntactic structure and the associated semantic roles of the arguments within those sentences. It delineates how verbs define the organisation of their associated arguments and the reflection of this organisation in a sentence's syntactic realisation. AST abstracts sentences as predicate-argument structures, where the predicate $p$ (associated with the verb) has a set of associated arguments $arg_i$, where each argument has an associated positional component $i$ and a thematic/semantic roles $r_i$, the latter categorising the semantic functions of arguments in relation to the verb (e.g. agent, patient, theme, instrument). In the context of this work, the AST predicate-argument representation is associated with a lexical-semantic representation of the content $c_i$ of the term $t_i$.

In this work, we simplify and particularise the relationship between the argument structure and the distributional lexical semantic representation as a \textit{role-content} relation, where the structural syntactic/semantic relationship is defined by its shallow semantics, i.e. as the composition of the content of the terms, their position in the predicate-argument (PArg) structure ($arg_i$) and their semantic roles (SRs) ($r_i$: $pred$, $arg$), as described below:
$$\underbrace{animals}_{ARG0}~\underbrace{require}_{PRED}~\underbrace{oxygen}_{ARG1}~\underbrace{for~survival}_{ARGM-PRP}$$
Therefore, we define the semantics of sentences, $sem(s)$, as the compositions between \textit{role-content}, which can be described as follows:
$
sem(s) = \underbrace{t_1({c_1}, {r_1})}_{i.e., ARG0-animals} \oplus \dots \oplus \underbrace{t_i({c_i}, {r_i})}_{PRP-survival}
$
Where $t_i({c_i}, {r_i})=c_i \otimes r_i$ represents the semantics of term $t_i$ with content $c_i$ (i.e., \textit{animals}) and SRL $r_i$ (i.e., \textit{ARG0}) in context $s$. $\otimes$: connects the meanings of words with their roles, using the compositional-distributional semantics notation of \cite{smolensky2006harmonic,Clark2007CombiningSA,clark2008compositional}. $\oplus$: connects the lexical semantics (word content + structural role) to form the sentence semantics. To deliver the localisation or composition property, the sentence semantics should be able to present separation or disentanglement under connector $\oplus$. E.g., replacing \textit{ARG0-animals} with \textit{ARG0-fishes}. 

\begin{figure}[t]
    \centering
    \includegraphics[width=\columnwidth]{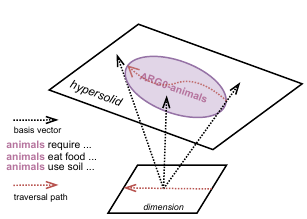}
    \caption{Algorithm \ref{alg:guide}: by modifying the latent dimensions, we can control the movement of latent vectors over latent space.}
    \label{fig:guide_trav}
\end{figure}

\paragraph{Formal semantic features in vector space.} After defining the semantic features of sentences, we propose the concept of a \textit{convex cone of semantic feature}. In linear algebra, a \textit{cone} refers to a subset of a vector space that is convex if any $\alpha \overrightarrow{v_i} + \beta \overrightarrow{v_j}$ if any $\overrightarrow{v_i}$ and $\overrightarrow{v_j}$ belong to it. $\alpha$ and $\beta$ are positive scalars. Formally, the definition of convex cone, $C$, is described as a set of vectors:
$
C = \{ x \in V | x = \sum_{i=1}^n \alpha_i v_i, \alpha_i \geq 0, v_i \in R \}
$
where $x$ is an element vector in vector space $\mathbb{R}$, $v_i$ are the basis vectors. $\alpha_i$ are non-negative scalars. In this context, we consider each \textit{role-content} feature as a convex cone, $C$, corresponding to a hypersolid in high-dimensional vector space:
$
C_{c_i, r_i} = \{ t({c_i}, {r_i}) | t({c_i}, {r_i}) \in sem(s), s \in \textit{corpus} \}
$
where $t({c_i}, {r_i})$ represents the basis vector in $C_{c_i, r_i}$ (Figure \ref{fig:guide_trav}). According to set theory, we can define the formal semantic space as follows:

\textit{\textbf{Assumption1:} The sentence semantic space is the union of all unique $C_{c_i, r_i}$ convex cones:}
$$
C_{c_1, r_1} \cup C_{c_2, r_2} \cup \dots \cup C_{c_{V^{(c)}}, r_{V^{(r)}}}
$$
$V$ is the vocabulary of a corpus. Based on Assumption1, we can establish:

\textit{\textbf{Proposition1:} The geometrical location of sentence semantic vectors, $sem(s)$, can be determined by the intersection of different $C_{c_i, r_i}$:}
\[
\begin{aligned}
    sem(s) & = t_1({c_1}, {r_1}) \oplus \dots \oplus t_i({c_{i}}, {r_{i}}) \\
    &= \{ t_1({c_1}, {r_1})\} \oplus \dots \oplus \{t_i({c_{i}}, {r_{i}}) \} \\
    &= C_{c_1, r_1} \cap C_{c_2, r_2} \cap \dots \cap C_{c_{i}, r_{i}}
\end{aligned}
\]
\section{Geometrical Formal Semantic Control}
In this section, we first show that our formal semantic geometry can interpret sentence generation, such as arithmetic \cite{shen2020educating}, and extend the ``Linear Representation Hypothesis''. Then, we propose a new semantic control approach, which recursively traverses the latent dimensions to probe the semantic geometry over latent spaces.

\paragraph{Geometrical algebra interpretability.} Arithmetic has been considered a common way to control word or sentence semantics over latent spaces \cite{mikolov-etal-2013-linguistic}. E.g., the addition operation can steer the sentence semantics \cite{shen2020educating,mercatali2021disentangling,liu2023context}, or linear interpolation can generate smooth intermediate sentences \cite{hu-etal-2022-fuse}. However, they lack an explanation for these phenomena. We show that our geometrical framework can provide an intuitive explanation for these phenomena. 

For linear interpolation, for example, it takes two sentences $ x_1 $ and $ x_2 $ and obtains latent vectors $ z_1 $ and $ z_2 $, respectively. It interpolates a path $ z_k = z_1 \cdot (1 - k) + z_2 \cdot k$  with $k$ increased from $ 0 $ to $ 1 $ by a step size of $ 0.1 $. 
Given two sentences with one role-content set overlap, $C_{c_{j}, r_{j}}$. We can describe:
\[
\begin{aligned}
    &sem(s_1) \cap sem(s_2) \\
    &=\{ C^{s_1}_{c_1, r_1} \cap \dots \cap C^{s_1}_{c_{i}, r_{i}} \} \cap \{C^{s_2}_{c_1, r_1} \cap \dots \cap C^{s_2}_{c_{i}, r_{i}} \} \\
    &=\{ C^{s_1}_{c_1, r_1} \cap \dots \cap C^{s_2}_{c_{i}, r_{i}} \} \cap C^{s_{1(2)}}_{c_{j}, r_{j}}
\end{aligned}
\]
According to the definition of convex cone, if $z_1$ and $z_2$ are left in $C^{s_{1(2)}}_{c_{j}, r_{j}}$, the weighted sum vector, $z_t$, is also in $C^{s_{1(2)}}_{c_{j}, r_{j}}$. Therefore, the intermediate sentence semantics can be described as:
\[
\begin{aligned}
    &sem(s^t_{1 \rightarrow 2}) \\
    &=(1-k) \times sem(s_1) + k \times sem(s_2) \\
    &=\{ \{z_1 \cdot (1 - k) + z_2 \cdot k \}, \dots \{\dots\}\} \cap C^{s_{1(2)}}_{c_{j}, r_{j}} \\
\end{aligned}
\]
That is, the intermediate sentences will hold the $\{c_{j}, r_{j}\}$ information during interpolation. 

\paragraph{Linear representation hypothesis.} ``Linear representation hypothesis'' refers to high-level concepts being represented linearly as directions in representation space, which has been widely evaluated to interpret Large LMs' mechanism \cite{marks2023geometry,xie2021explanation,wang2024large,jiang2024origins,park2023linear,park2024geometry}. However, a main challenge for this hypothesis is that it’s not clear what constitutes a high-level concept.

Our geometrical framework can further support and extend this hypothesis by answering the questions: What and how are they ``linearly'' encoded? For example, given a set of $N$ atomic sentences: $s_i$: \textit{bird is a kind of living thing} varying the content of arg1. Their semantics can be described below:
\[
\begin{aligned}
& sem(s) =  \{ C^{s_i}_{c_i, arg1}, \dots \} \cap \dots \cap C_{living~ thing, arg2} \\
&, \text{where}~ c_i \in \{ \text{tiger}, \text{bird}, \dots \}
\end{aligned}
\]
In this case, the concept \textit{living thing} is encoded as a convex cone where all different $C^{s_i}_{c_i, arg1}$ contribute to its boundary, leading to a direction. The hierarchical relations between \textit{living thing} and \textit{bird, etc.} are determined by the convex cones \textit{is a kind of}.

\paragraph{Guided traversal.} Since we describe different sentence semantic features, $\{c_i, r_i\}$, as distinct convex cones, $C_{c_i, r_i}$, within a $N$-dimensional vector space, $V \in \mathbb{R}^{N}$, we can linearly divide each basis dimension, $i \in {\{1, \dots, N \}}$, into different value regions, $[a, b]^{(i)}$, based on minimal information entropy. Consequently, there is a sequence of dimensional subspaces for each semantic feature. Thus, movement between different $C_{c_i, r_i}$ regions can be achieved by moving out the dimensional regions within this sequence. This process can be implemented via a decision tree. In figure \ref{fig:decision_tree}, for example, we can move the sentence from $C_{pred, causes}$ to $C_{pred, means}$ by modifying the values started from \textit{dim 21 $\le -0.035$}, ..., ending at \textit{dim 10 $\le -1.11$}. By traversing the tree path, we can control the sentence generation by moving between convex cones, detailed in Algorithm \ref{alg:guide}.
\begin{algorithm}[ht!]
\caption{Guided latent space traversal} \label{alg:guide}
\begin{algorithmic}[1]
\State Datasets: $D = \{s_1, \dots, s_n\}$ 
\State Labels: $Y = \{y_1, \dots, y_n \}$, $y_i \in \{0, 1\}$ 
\State \textit{\# \textcolor{blue}{0:pred-causes}, \textcolor{red}{1:pred-means}}
\State Seed: $s=$ \textit{fire \textcolor{blue}{causes} chemical change}
\For{$s_i \in D$}
\State $z_i \leftarrow \text{Encoder}(s_i)$
\EndFor
\State $X \leftarrow \{z_1, \dots, z_n\}$
\State $\text{tree} \leftarrow \text{DecisionTreeClassifier}(X,Y)$
\State $\text{path} \leftarrow \text{filter}(\text{tree})$ \textit{\# choose the shortest path between $C_0$ and $C_1$}
\State $z \leftarrow \text{Encoder}(s)$
\For{$\text{node} \in \text{path}$}
    \State (dim, range, yes/no) $\leftarrow$ node 
    \State \textbf{if} in current branch \textbf{do} 
    \State ~ z[dim] $\leftarrow$ $v \notin \text{range}$ \textbf{if} yes \textbf{else} $v \in \text{range}$
    \State \textbf{else do} 
    \State ~ z[dim] $\leftarrow$ $v \in \text{range}$ \textbf{if} yes \textbf{else} $v \notin \text{range}$
    
\EndFor
\State $s \leftarrow$ \text{Decoder}(z) \textit{\# fire \textcolor{red}{means} chemical change}
\end{algorithmic}
\end{algorithm}
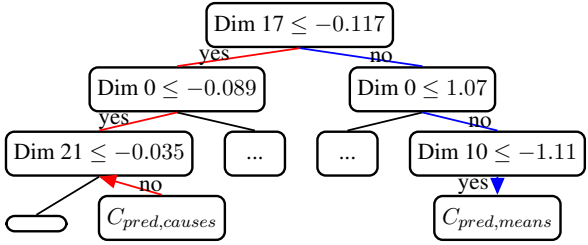
\begin{figure}[ht!]
\centering
\resizebox{7.9cm}{!}{
\begin{forest}
for tree={draw, rounded corners, node options={align=center, minimum width=1cm, line width=0.4mm}, s sep=5mm, l sep=3mm, edge={line width=0.3mm, - }}
[Dim 17 $\leq -0.117$,edge+={draw=blue},edge+={draw=red}
    [Dim 0 $\leq -0.089$,edge+={draw=red},edge+={draw=red}, edge label={node[midway, left, draw=none]{yes}}
        [Dim 21 $\leq -0.035$,edge+={draw=red}, edge label={node[midway, left, draw=none]{yes}}
            []
            [$C_{pred,causes}$, edge+={<-, draw=red}, edge label={node[midway, right, draw=none]{no}}]
        ]
        [...]
    ]
    [Dim 0 $\leq 1.07$,edge+={draw=blue},edge label={node[midway, right, draw=none]{no}}
        [...]
        [Dim 10 $\leq -1.11$,edge+={draw=blue}, edge label={node[midway, right, draw=none]{no}}
            [$C_{pred,means}$, edge+={->, draw=blue}, edge label={node[midway, left, draw=none]{yes}}]]
    ]
]
\end{forest}
}\caption{Traversal between different role-content sets by moving along the tree path.}
\label{fig:decision_tree}
\end{figure}

Based on our algorithm, we can use classification metrics as proxy metrics to evaluate latent space geometry. E.g., accuracy and recall for measuring feature \textit{separability} and \textit{density}. 
\section{SRL-Conditional VAE} \label{sec:disentang} 
In this section, we investigate the architecture of VAE to integrate the latent sentence space with LMs and propose a supervision approach to learn formal semantic geometry (i.e., role-content).
\paragraph{Model architecture.} We consider Optimus \cite{li2020optimus} as the foundation which used BERT and GPT2 as Encoder and Decoder, respectively. In detail, the sentence representation, $\text{Embed(x)}$, encoded from BERT[cls] will first transform into a Gaussian space by learning the parameters $\mu$ and $\sigma$ through multilayer perceptions $W_\mu$, $W_{\sigma}$. The final latent sentence representations can be obtained via: $z = W_{\mu} \times \text{Embed(x)} + W_{\sigma}$, which, as an additional Key and Value, is concatenated into the original Key and Value weights of GPT2, which can be described as:
$
\text{Attention}(Q, K, V) = \text{softmax}( \frac{Q [z;K]^T}{\sqrt{d}})[z;V]
$
where $Q$ has the shape $\mathbb{R}^{\text{seq} \times 64}$, $K, V$ has the shape $\mathbb{R}^{(\text{seq}+1) \times 64}$ (64 is dimension of GPT2 attention, $\text{seq}$ is sequence length). Since $Q$ represents the target, $K$ and $V$ represent the latent representations. By intervening the $KV$ with $z$, we can learn the transformation between latent space and observation distribution. 

\paragraph{Optimisation.} It can be trained via the evidence lower bound (ELBO) on the log-likelihood of the data $x$ \cite{Kingma2014AutoEncodingVB}. To bind the word content and semantic role information in latent space, we conditionally inject the semantic role sequence into latent spaces where the latent space $z$ and semantic role $r$ are dependent. The joint distribution can be described as:
$$
P_{\theta}(x,r,z)=\underbrace{P_{\theta}(x|z,r)}_{likelihood} \times \underbrace{P_{\theta}(z|r)}_{prior} \times P(r)
$$
\begin{figure}[ht!]
\centering
\resizebox{7.8cm}{!}{
\begin{minipage}{3.5cm}
\begin{tikzpicture}
  \node[latent] (r) at (2.4,2) {\tiny pregroup};
  \node[latent] (z) at (0,2) {\tiny tokens};
  \node[latent] (x) at (1.2,0.15) {\tiny language};
  \edge {z, r} {x};
  \path (z) edge[->] node[pos=0.5,below,sloped] {\small \textit{compose}} (x);
  \path (r) edge[->] node[pos=0.4,below,sloped] {\small \textit{type logic}} (x);
\end{tikzpicture}
\subcaption{CDM \cite{clark2008compositional}}
\end{minipage}
\hspace{1cm}
\begin{minipage}{3.5cm}
\begin{tikzpicture}
  \node[latent] (r) at (2.4,2) {$r$};
  \node[latent] (z) at (0,2) {$z$};
  \node[latent] (x) at (1.2,0) {$x$};
  \edge {r} {z};
  \edge {z,r} {x};
  \path (r) edge[->] node[below] {$p_\theta(z|r)$} (z);
  \path (z) edge[->] node[below,sloped] {\small $p_\theta(x|z,r)$} (x);
\end{tikzpicture}
\subcaption{SRL-Conditional VAE}
\end{minipage}
\hspace*{1pt}
}
\caption{Comparison between Compositional Distributional Model (CDM) (left) and SRL-Conditional VAE (right).}
\label{fig:cvae}
\end{figure}
Specifically, we first model the categorical structures by encoding the semantic roles sequence to learn the prior distribution with parameters $\mu^{(srl)}$ and $\sigma^{(srl)}$. Then, we jointly encode semantic roles and lexical tokens to learn the approximate posterior parameterised by $\mu$ and $\sigma$. By minimising the Kullback-Leibler (KL) divergence between prior and approximate posterior, the semantic features can be encoded in the latent sentence space.
Moreover, to avoid the KL vanishing problem, which refers to the KL term in the ELBO becomes very small or approaching zero, we select the cyclical schedule to increase weights of KL $\beta$ from 0 to 1 \cite{fu-etal-2019-cyclical} and a KL thresholding scheme \cite{li-etal-2019-surprisingly} that chooses the maximum between KL and threshold $\lambda$. The final objective function can be described as follows:
\begin{align*} 
\mathcal{L}_\text{CVAE} = & - \mathbb{E}_{q_\phi(z|r,x)} \Big[ \log p_{\theta} ( x | z,r ) \Big]  \\ \nonumber
& + \beta \sum_i \max \left[ \lambda , \text{KL} q_\phi(z_i|x, r) || p(z_i|r) \right ]
\end{align*}
where $q_\phi$ represents the approximated posterior (i.e., encoder). $i$ is the $i$-th latent dimension.



\section{Empirical analysis} \label{sec:empirical}
In the experiment, we quantitatively and qualitatively evaluate the latent space geometry via geometrical probing approaches: (1) traversal, (2) arithmetic, and (3) guided traversal. All experimental details are provided in Appendix \ref{sec:dsr_labels}.

\subsection{Latent Traversal}
\paragraph{Qualitative evaluation.} Traversal refers to the random walk over latent space. It can be done by decoding the latent vector in which each dimension is resampled and other dimensions are fixed \cite{higgins2016beta,kim2018disentangling,carvalho2022learning}. Given a latent vector from a ``seed'' sentence, we can traverse its neighbours to evaluate the geometry. As illustrated in Table~\ref{tab:trav_examples}, those traversed sentences can hold the same content under different semantic roles as the input, such as \textit{automobile} in \textit{ARG1}, indicating \textit{role-content} feature separation in latent spaces.

\begin{figure}[ht!]
    \includegraphics[width=\columnwidth]{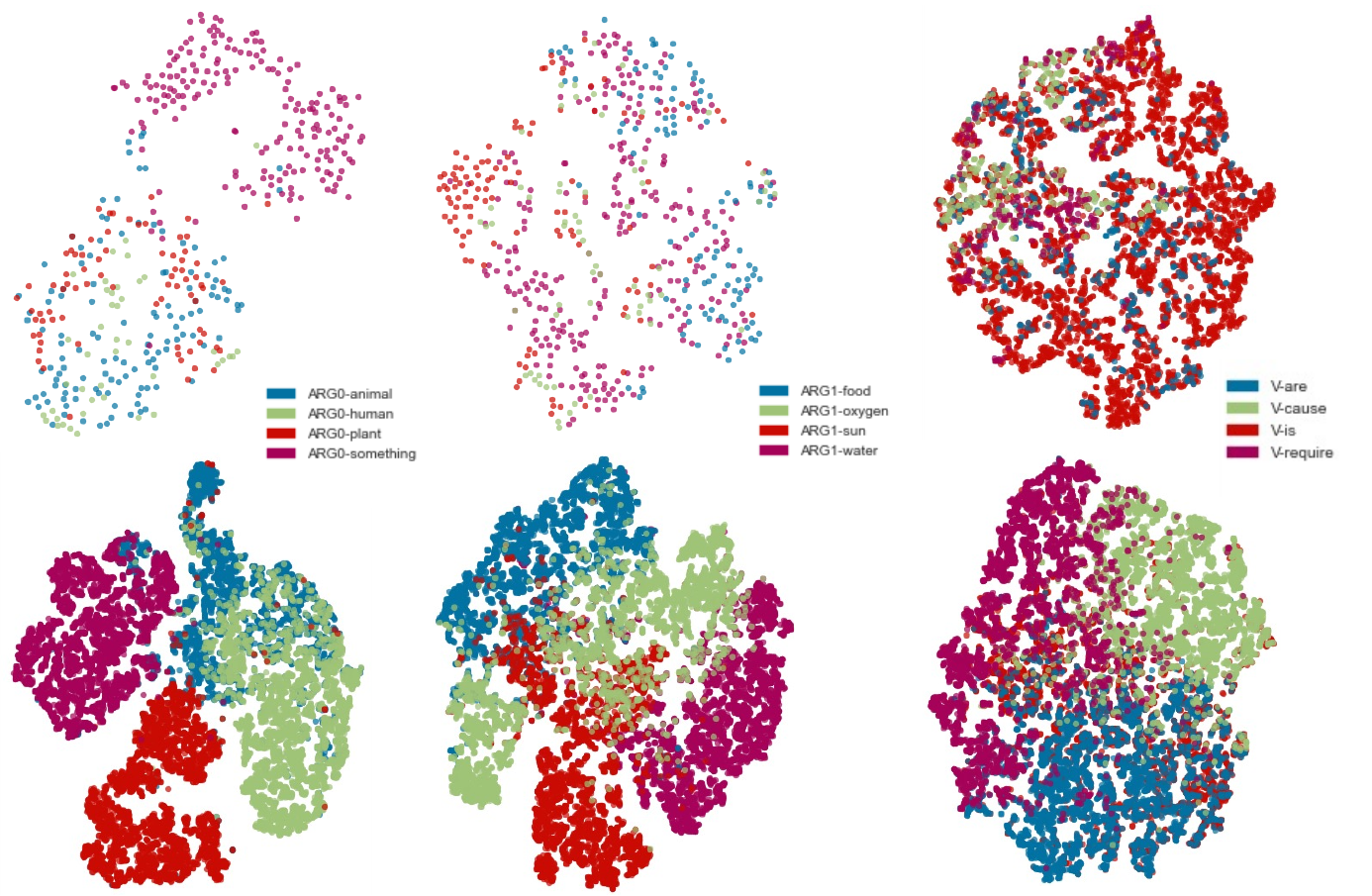}
    \caption{t-SNE plot of role-content distribution before and after traversal. From left to right are ARG0-(animal, human, plant, and something), ARG1-(food, oxygen, sun, and water), and predicate-(are, cause, is, require) (top: original role-cluster distribution, bottom: distribution after traversal). PCA plots are in Figure \ref{fig:pca}.}
    \label{fig:da_arith_cluster}
\end{figure}

%
%
\begin{table}[h]




\begin{tcolorbox}[fontupper=\small, fontlower=\small, middle=0.3cm, top=1pt, bottom=1pt]
\underline{an automobile is a kind of vehicle} \\ \\
\textcolor{blue}{an automobile} is a kind of moving object  \\
\textcolor{blue}{an automobile} is a kind of object \\
\\
an airplane is \textcolor{blue}{a kind of vehicle} \\
a car is \textcolor{blue}{a kind of vehicle}


\end{tcolorbox}
\caption{Traversal showing \textcolor{blue}{held} semantic factors in explanations corpus.}
\label{tab:trav_examples}
\end{table}
\paragraph{Quantitative evaluation.} Next, we employ t-SNE \cite{van2008visualizing} to examine \textit{role-content} features cluster and separation over latent space (i.e., \textit{natural clustering property} \cite{bengio2013deep}). In the corpus, however, due to the small number of data points within each role-content cluster, t-SNE cannot capture the differences between clusters well, resulting in the visualized latent space not displaying good role-content separability (top in figure \ref{fig:da_arith_cluster}). Therefore, we increase the number of data points in different role-content clusters by traversing each and keeping those resulting data points with the same role-content. Then, we visualise the role-content cluster at the bottom of figure \ref{fig:da_arith_cluster}. We can find that the features are clustered and separated over the latent space. If this was not the case, after traversing the resulting vectors from the same role-content cluster, the visualization should show the same entanglement as the original datapoints distribution.
\subsection{Latent Arithmetic}
\paragraph{Qualitative evaluation.} In addition, we demonstrate the geometric properties via interpolation in Table~\ref{tab:interpol_examples}. 
\begin{table}[ht!]
\begin{tcolorbox}[fontupper=\small, fontlower=\small, top=1pt, bottom=1pt]
\textcolor{blue}{a beach ball is a kind of container} \\
1. a pool table is a kind of object \\
2. a balloon is a kind of object \\
3. a magnet is a kind of object \\
4. a neutron is a kind of particle \\
5. a proton is a kind of particle\\
\textcolor{blue}{an atom is a kind of particle}
\tcblower
\textcolor{blue}{protons are found in the nucleus of an atom} \\
1. protons are found in the nucleus of an atom \\
2. 1 atom is positive 1 in electric charge \\
3. \textcolor{red}{1 in 6000 is equal to 27 in 10 years} \\ 
4. if protons and neutrons have the same number of neutrons then those two particles are physically closer than one another \\
5. if a neutron has a negative -10 electric charge then the atom will not be able to move \\
6. if a neutron has a negative -10 electric charge then the neutron will not have a positive electric charge \\
\textcolor{blue}{if a neutral atom loses an electron then an atom with a positive charge will be formed}
\end{tcolorbox}
\caption{Interpolation examples (top: interpolation between sentences with similar semantic information, bottom: interpolation between sentences with different semantic information). Only unique sentences shown.}
\label{tab:interpol_examples}
\end{table}
For the top-most one, we can observe that sentences are smoothly moved from source to target (e.g., from \textit{beach ball} to \textit{atom} connected by \textit{ballon}, \textit{magnet}, \textit{neutron}, and \textit{proton}) where the same role-content (i.e., \textit{pred-is}) unchanged. In contrast, the second case doesn't display the smooth interpolation path. E.g., the third sentence connecting different semantic structures is unrelated to both source and target due to a discontinuous space gap between different clusters. Both indicate that the explanatory sentences might be clustered according to different semantic role structures.

\begin{table}[ht]
\begin{tcolorbox}[fontupper=\small, fontlower=\small, middle=0.3cm, top=1pt, bottom=1pt]
\underline{$s_1$: animals require food for survival} \\
\underline{$s_2$: animals require warmth for survival} \\
\textcolor{blue}{animals} eat plants  \\
\textcolor{blue}{animals} produce milk \\
\textcolor{blue}{animals} usually eat plants \\
\textcolor{blue}{animals} eat berries ; plants \\
\textcolor{blue}{animals} require food to survive \\
\textcolor{blue}{animals} require shelter to survive
\tcblower
\underline{$s_1$: water vapor is invisible} \\
\underline{$s_2$: the water is warm} \\
\textcolor{blue}{igneous rocks} are found under the soil \\
\textcolor{blue}{quartz} is usually very small in size \\
\textcolor{blue}{quartz} is formed by magma cooling \\
\textcolor{blue}{quartz} is made of iron and zinc \\
\textcolor{blue}{silica} is made of argon and argon \\
\textcolor{blue}{sedimentary} is formed by lithosphere collapsing
\end{tcolorbox}
\caption{$s_1 \pm s_2$ (top: addition, bottom: subtraction).}
\label{tab:arith_examples}
\end{table}

Following the definition of convex cone, we next traverse the resulting sentence after adding or subtracting two sentences with the same role-content feature. As illustrated in Table~\ref{tab:arith_examples}, the adding operation tends to hold the same role-content (e.g., \textit{ARG0-Animals}) as inputs. In contrast, the subtraction loses such control, e.g., from \textit{ARG1-water} to \textit{ARG1-quartz}.   
More similar observations are in Table \ref{tab:arith_other_examples}. These results corroborate our geometry.

\begin{figure}[ht!]
    \centering
    \includegraphics[scale=0.36]{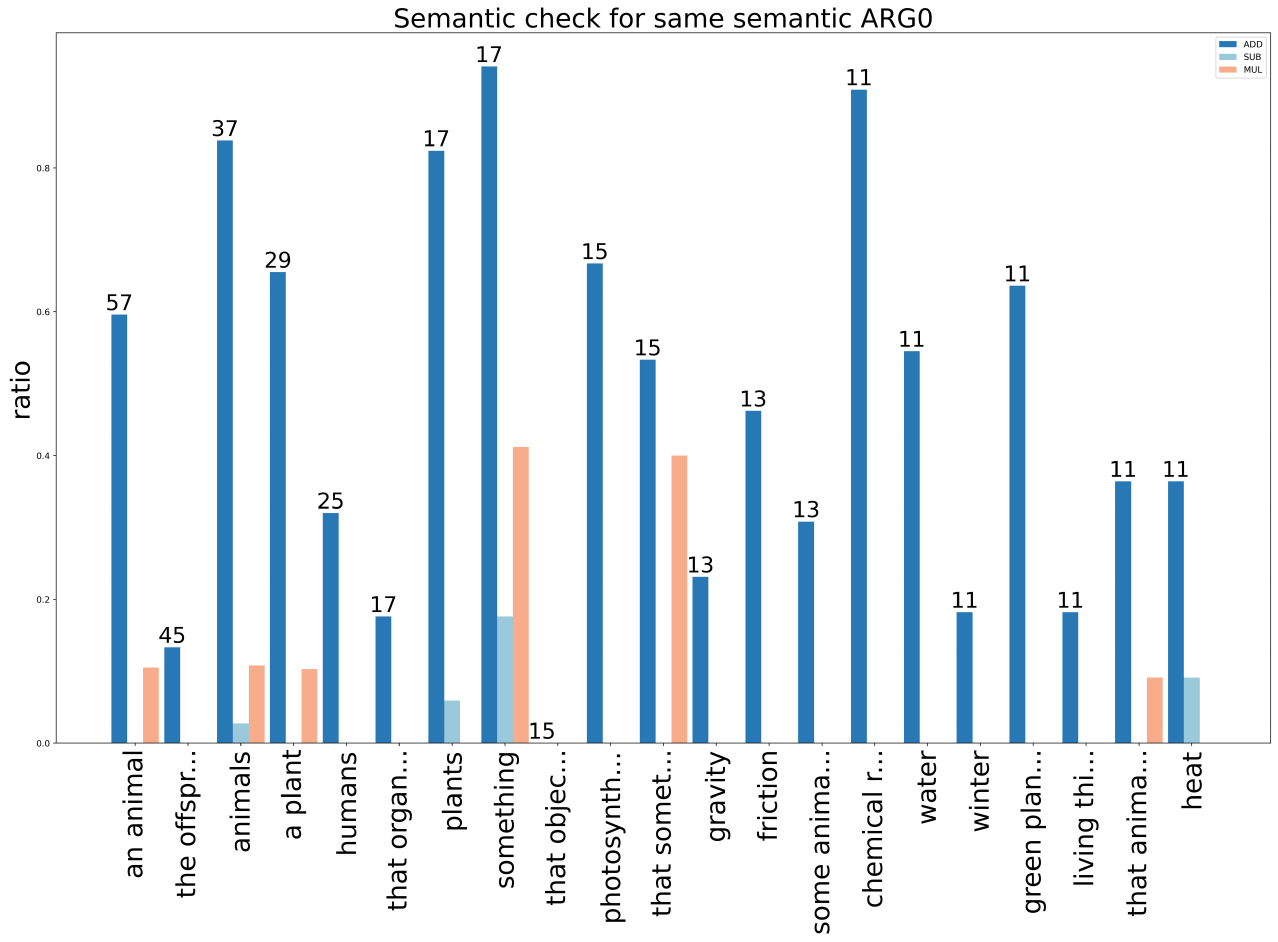}
    \caption{Arithmetic, $s_1 \pm s_2$, for ARG0 with contents (dark blue: addition, shallow blue: subtraction, orange: element-wise production).}
    \label{fig:a0_animal}
\end{figure}

\begin{figure}[ht!]
    \centering
    \includegraphics[scale=0.35]{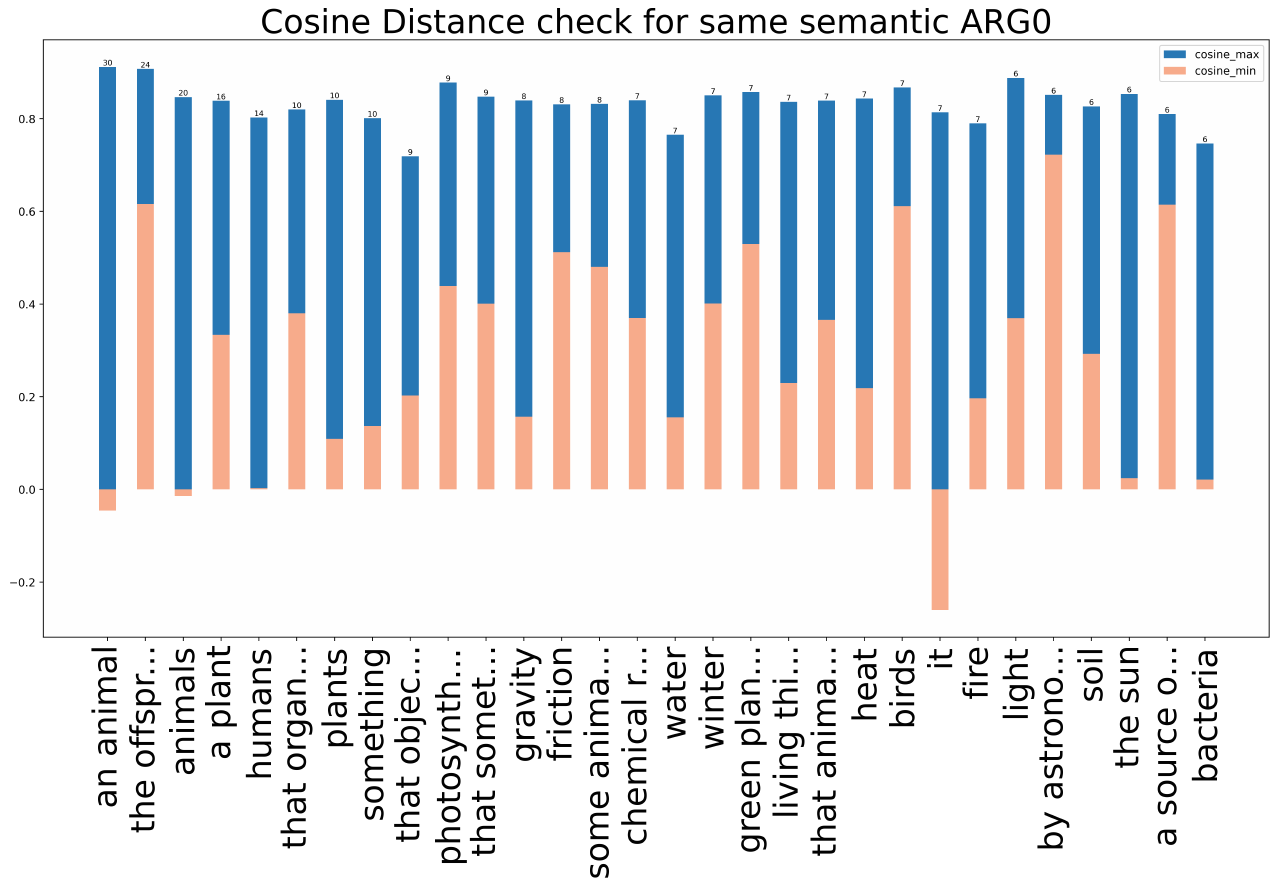}
    \caption{Evaluating the geometrical size of role-content clusters (blue: max, orange: min).}
    \label{fig:a0_size}
\end{figure}

\paragraph{Quantitative evaluation.} Next, we quantitatively assess our geometry framework by calculating the ratio of the same role-content results from the vector addition and subtraction for all sentence pairs with a matching role. As illustrated in Figure \ref{fig:a0_animal}, the ADDed results (dark blue) can greatly hold the same token-level semantics (role-content) as inputs, indicating our geometrical framework. In contrast, the SUBed results (shallow blue) suffer from semantic shift. Similar observations for VERB and ARG1 can be found in Figure \ref{fig:consistency_verb_sem} and \ref{fig:consistency_arg1_sem}.
Besides, we can quantify each role-content cluster's geometrical area by calculating the cosine similarity between randomly selected sentence pairs in this cluster. We report the maximal and minimal distance in Figure \ref{fig:a0_size}. Similar observations for VERB and ARG1 can be found in Figure \ref{fig:cos_dist_v} and \ref{fig:cos_dist_a1}.

\subsection{Guided Latent Traversal} \label{sec:guide}


Finally, we examine the semantic geometry via algorithm\ref{alg:guide}. The categories selected below are chosen based on their frequencies, ensuring a balanced distribution during the classifier's training process.

\paragraph{Qualitative evaluation.} Firstly, we evaluate the traversal between different semantic role structures, e.g, conditional and atomic sentences. Table \ref{tab:movement_topic_example} shows that the cluster of the generated sentence changes as the values of different dimensions change sequentially (e.g., the first three sentences hold the same characteristic \textit{if ... then ...} as the input. The remaining sentences gradually move closer to the target characteristics, such as \textit{is}). Meanwhile, the sentences can hold the subject, \textit{something}, during the movement, corroborating our geometry framework.
\begin{table}[ht!]
\begin{tcolorbox}[fontupper=\small, fontlower=\small, middle=0.3cm, top=1pt, bottom=1pt]

\ul{if something receives sunlight it will absorb the sunlight} \\
Dim27: \textcolor{blue}{if} a thing absorbs sunlight \textcolor{blue}{then} that thing is warmer  \\
Dim12: \textcolor{blue}{if} something is eaten \textcolor{blue}{then} that something produces heat \\
Dim08: \textcolor{blue}{if} something gets too hot in sunlight \textcolor{blue}{then} that something is less able to survive \\
Dim03: something \textcolor{blue}{contains} physical and chemical energy \\
Dim21: something \textcolor{blue}{contains} sunlight \\
Dim10: some things \textcolor{blue}{are} made of matter \\
Dim00: something \textcolor{blue}{is} made of atoms \\
Dim17: \textcolor{red}{a forest} \textcolor{blue}{contains} life \\
Dim00: something that is cold \textcolor{blue}{has} a lower temperature \\
Dim21: something \textcolor{blue}{rises} in temperature \\
Dim00: something \textcolor{blue}{is} formed from things dissolved in water \\
Dim30: something that is cold \textcolor{blue}{has} fewer nutrients \\
Dim21: something that is not moved \textcolor{blue}{is} dead
\end{tcolorbox}
\caption{Movement from \textit{conditional} to \textit{atomic} sentences.}
\label{tab:movement_topic_example}
\end{table}
Next, we evaluate the traversal between predicates. Table \ref{tab:movement_verb_example} shows the movement between verbs (\textit{cause} and \textit{mean}). We can observe that the predicate is modified from \textit{causes} to \textit{mean}. In the traversal process, some sentences fall into the \textit{V-is} region. The reason is that the \textit{V-is} cluster is widely scattered in latent space (shown in Figure \ref{fig:da_arith_cluster}), which leads to a big overlap between \textit{V-is} and \textit{V-mean}. Moreover, we calculate the ratio of the generated sentences that hold the expected predicate, \textit{mean}, from 100 sentences with predicate \textit{cause}. The ratio is 0.71, which indicates that the decision tree is a reliable way to navigate the movement of sentences.
\begin{table}[ht!]
\begin{tcolorbox}[fontupper=\small, fontlower=\small, middle=0.3cm, top=1pt, bottom=1pt]
\ul{fire causes chemical change} \\
Dim06: fire \textcolor{blue}{causes} chemical changes \\
Dim22: fire \textcolor{blue}{causes} chemical reactions \\
Dim02: fire can \textcolor{blue}{cause} harm to plants \\
Dim27: smoke can \textcolor{blue}{cause} harm to organisms \\
Dim14: fire \textcolor{blue}{causes} physical harm to objects \\
Dim24: fire can \textcolor{blue}{cause} chemical changes \\
Dim08: fire \textcolor{blue}{destroys} material \\
Dim01: fire \textcolor{blue}{means} chemical change \\
Dim14: \textcolor{red}{waste} \textcolor{blue}{means} igneous metal \\
Dim06: \textcolor{red}{combustion} \textcolor{blue}{means} burning \\
Dim00: \textcolor{red}{combustion} \textcolor{blue}{means} chemical changes \\
Dim21: \textcolor{red}{combustion} \textcolor{blue}{means} burning \\
Dim00: fire \textcolor{blue}{is} formed by thermal expansion \\
Dim18: fire chemical \textcolor{blue}{means} chemical energy \\
Dim03: fire \textcolor{blue}{is} corrosive
\tcblower
\ul{winter means cold environmental temperature}\\
Dim03: winter \textcolor{blue}{means} cold - weather\\
Dim18: winter \textcolor{blue}{means} cold weather \\
Dim00: winter \textcolor{blue}{means} weathering \\
Dim21: \textcolor{red}{drought} \textcolor{blue}{means} high temperatures / low precipitation \\
Dim00: winter \textcolor{blue}{means} high amounts of precipitation \\
Dim06: \textcolor{red}{drought} \textcolor{blue}{causes} natural disasters \\
Dim14: \textcolor{red}{drought} \textcolor{blue}{has a negative impact on} crops \\
Dim01: \textcolor{red}{drought} \textcolor{blue}{has a negative impact on} animals \\
Dim08: \textcolor{red}{drought} \textcolor{blue}{causes} animal populations to decrease \\
Dim24: \textcolor{red}{drought} \textcolor{blue}{causes} ecosystem loss \\
Dim14: \textcolor{red}{drought} \textcolor{blue}{causes} animals to have lower natural temperature \\
Dim27: cold climates \textcolor{blue}{causes} wildfires \\
Dim02: \textcolor{red}{climate change} can \textcolor{blue}{cause} low rainfall \\
Dim22: \textcolor{red}{global warming} \textcolor{blue}{causes} droughts \\
Dim06: winter \textcolor{blue}{causes} weather patterns
\end{tcolorbox}
\caption{Movement between \textit{cause} and \textit{mean}.}
\label{tab:movement_verb_example}
\end{table}
Finally, we evaluate the traversal between arguments. Table \ref{tab:movement_arg_example} shows the movement from argument \textit{water} to \textit{something}. Similarly, the ARG1 can be modified from \textit{water} to \textit{something} following its path. Besides, the final generated explanation still holds a similar semantic structure, \textit{is a kind of}, compared with input.
\begin{table}[ht!]
\begin{tcolorbox}[fontupper=\small, fontlower=\small, middle=0.3cm, top=1pt, bottom=1pt]

\ul{water is a kind of substance} \\
Dim12: \textcolor{blue}{water} is a kind of substance  \\
Dim00: \textcolor{blue}{water} is a kind of liquid \\
Dim23: \textcolor{blue}{liquid} is a kind of material \\
Dim29: \textcolor{blue}{water} has a positive impact on a process \\
Dim17: absorbing \textcolor{blue}{water} is similar to settling \\
Dim06: \textcolor{red}{absorbing} is similar to reducing \\
Dim21: absorbing \textcolor{blue}{something} is similar to absorbing something \\
Dim04: storing \textcolor{blue}{something} means being protected \\
Dim06: producing \textcolor{blue}{something} is a kind of process \\
Dim04: storing \textcolor{blue}{something} is similar to recycling \\
Dim21: absorbing \textcolor{blue}{something} is a kind of process \\
Dim01: absorbing \textcolor{blue}{something} can mean having that something \\
Dim22: folding \textcolor{blue}{something} is similar to combining something \\
Dim07: improving \textcolor{blue}{something} is a kind of transformation \\
Dim11: absorbing \textcolor{blue}{something} is a kind of method \\
Dim07: absorbing \textcolor{blue}{something} is a kind of process
\end{tcolorbox}
\caption{Movement from \textit{water} to \textit{something}.}
\label{tab:movement_arg_example}
\end{table}

\paragraph{Quantitative evaluation.} Finally, we use classification metrics, including accuracy (\textit{separability}) and recall (\textit{density}), as proxy metrics to assess latent space geometry. As shown in Table \ref{tab:proxy_metrics}, all features show higher separation where argument1 leads to the highest separation, indicating latent space geometry. 




\begin{table}[ht!]
    \centering
\resizebox{7.8cm}{!}{
\begin{tabular}{lll}  \toprule
    \textbf{Formal semantic features}  & \textbf{separation}$\uparrow$ & \textbf{density}$\uparrow$  \\ \hline
    predicate (causes, means) & 0.87 & 0.92 \\
    argument1 (water, something) & 0.95 & 0.48 \\
    structure (condition, atomic) & 0.58 & 0.55 \\ \toprule
\end{tabular}
}
    \caption{Proxy metrics for latent space geometry.} 
    \label{tab:proxy_metrics}
\end{table}



\section{Conclusion and Future Work} \label{sec:concl}
In this study, we investigate the localisation of general semantic features to enhance the controllability and explainability of distributional space from the perspective of formal semantics, which is currently under-explored in the NLP domain. We first propose the formal semantic features as \textit{role-content} and define the corresponding geometrical framework. Then, we propose a supervision approach to bind the semantic role and word content. In addition, we propose a novel traversal probing approach to assess the latent space geometry based on information set and entropy. We extensively evaluate the latent space geometry through geometrical operations, such as traversal, arithmetic, and our guided traversal. Experimental results indicate the existence of formal semantic geometry.

Since recent theoretical works reveal that the LLMs can encode linear symbolic concepts \cite{jiang2024origins}, 
in the future, we will explore their in-context learning of compositional semantics based on our formal semantic geometry framework.

\section{Limitations}
\textbf{1.} Limitation of data source: this work only focused on explanatory sentences. Whether the semantic separability of other corpora emerges over latent space is not explored. \textbf{2.} Role-content clusters overlapping: the geometric analysis indicates that the role-content regions still have significant overlapping over distributional spaces. Therefore, a new potential task can be how we can better separate/disentangle the semantic features (role-content) to provide better localisation or composition behaviour over distributional semantic spaces in the Computational Linguistics domain, further assisting downstream tasks, such as Natural Language Reasoning, Compositional Generalisation, etc. \textbf{3.} Large Language Models: this paper only investigates the BERT-GPT2 architecture based on the current state-of-the-art language VAE (Optimus). The larger decoder is out of the scope of this work and needs to be investigated in the future.

\section*{Acknowledgements}
We appreciate the reviewers for their insightful comments and suggestions. This work was partially funded by the EPSRC grant EP/T026995/1 entitled “EnnCore: End-to-End Conceptual Guarding of Neural Architectures” under Security for all in an AI enabled society, by the Swiss National Science Foundation (SNSF) project NeuMath (\href{https://data.snf.ch/grants/grant/204617}{200021\_204617}), by the CRUK National Biomarker Centre, and supported by the Manchester Experimental Cancer Medicine Centre and the NIHR Manchester Biomedical Research Centre.

\bibliography{references}
\bibliographystyle{acl_natbib}

\appendix
\clearpage
\appendix

\section{Experiment Setting} \label{sec:dsr_labels}
\paragraph{Dataset.} Table \ref{tab:stats_data} displays the statistical information of the datasets used in the experiment. The data of the two datasets partially overlap, so only the unique explanations are selected as the experimental data. The rationale for choosing explanatory sentences is that they are designed for formal/localised/symbolic semantic inference task in natural language form, which provides a semantically complex and yet controlled experimental setting, containing a both well-scoped and diverse set of target ``concepts'' and sentence structures, providing a semantically challenging yet sufficiently well-scoped scenario to evaluate the syntactic and semantic organisation of the space. Besides, those concepts mentioned in the corpus, such as \textit{animal is a kind of living thing}, are fundamental to human semantic understanding.
\begin{table}[ht!]
    \small
    \centering
    \renewcommand\arraystretch{1.3}
      \resizebox{7.6cm}{!}{
    \begin{tabular}{|c|cc|}
        \hline
        Corpus & Num data. & Avg. length \\ \hline
        WorldTree \cite{jansen2018worldtree} & 11430 & 8.65 \\
       EntailmentBank \cite{https://doi.org/10.48550/arxiv.2104.08661} & 5134 & 10.35 \\ \hline
    \end{tabular}
    }
    \caption{Statistics from explanations datasets.} \label{tab:stats_data}
\end{table}

Table \ref{tab:visua_details} illustrates the semantic, structure, and topic information of explanatory sentences over the latent space. The explanatory sentences are automatically annotated using the semantic role labelling (SRL) tool, which can be implemented via AllenNLP library \cite{Gardner2017ADS}. We report in Table~\ref{tab:srl_silva} the semantic roles from the explanations corpus. 
\begin{table*}[ht!]
    \small \setlength\tabcolsep{4.5pt}
    \centering
\renewcommand\arraystretch{1.1}
    \begin{tabular}{|p{1cm}|p{14cm}|}  \hline
        \textbf{Cluster}         & \textbf{Theme and Pattern}                       \\ \hline
        0 & Theme: physics and chemistry. Pattern: \textit{if then} and \textit{as}. E.g., if a substance is mixed with another substance then those substances will undergo physical change.   \\ \hline
        1 & Theme: country, astronomy, and weather. E.g., new york state is on earth \\ \hline
        2 & Theme: physics and chemistry. Pattern: \textit{is a kind of}. E.g., light is a kind of wave. \\ \hline
        3 & Theme: biology. E.g., a mother births offspring. \\ \hline
        4 & Theme: synonym for verb. Pattern: \textit{means} and \textit{is similar to}. E.g., to report means to show. \\ \hline
        5 & Theme: astronomy. E.g., the solar system contains asteroids.\\ \hline
        6 & Theme: animal/plant. Pattern: \textit{is a kind of}. E.g., a seed is a part of a plant. \\ \hline
        7 & Theme: item. E.g., a telephone is a kind of electrical device for communication.\\ \hline
        8 & Theme: synonym for life. Pattern: \textit{means} and \textit{is similar to}. E.g., shape is a kind of characteristic.  \\ \hline
        9 & Theme: geography. Pattern: \textit{is a kind of}. E.g., a mountain is a kind of environment.\\ \hline
        10 & Theme: animal and plant. Pattern: \textit{if then} and \textit{as}. E.g., if a habitat is removed then that habitat is destroyed.  \\ \hline
        11 & Theme: scientific knowledge. Pattern: \textit{(;)}, \textit{number} and \textit{/}. E.g., freezing point is a property of a ( substance ; material ). \\ \hline
        12 & Theme: item. Pattern: \textit{is a kind of object}. E.g., a paper is a kind of object. \\ \hline
        13 & Theme: chemistry and astronomy. E.g., oxygen gas is made of only oxygen element.\\ \hline
        14 & Theme: general about science. Pattern: \textit{(;)}. E.g., seed dispersal has a positive impact on ( a plant ; a plant 's reproduction). \\ \hline
        15 & Theme: item. Pattern: \textit{is a kind of}. E.g., fertilizer is a kind of substance. \\ \hline
        16 & Theme: physics and chemistry. Pattern: \textit{(;)}. E.g., the melting point of oxygen is -3618f ; -2188c ; 544k. \\ \hline
        17 & Theme: animal. E.g., squirrels live in forests. \\ \hline
        18 & Theme: nature. E.g., warm ocean currents move to cooler ocean regions by convection.\\ \hline
        19 & Theme: life. E.g., pond water contains microscopic living organisms.\\ \hline
    \end{tabular}
    \caption{Cluster Information for explanatory sentences, we use a k-means classifier to classify the sentence representations and manually evaluate each class.} 
    \label{tab:visua_details}
\end{table*}

\begin{figure}[ht!]
\begin{center}
    \includegraphics[width=\columnwidth]{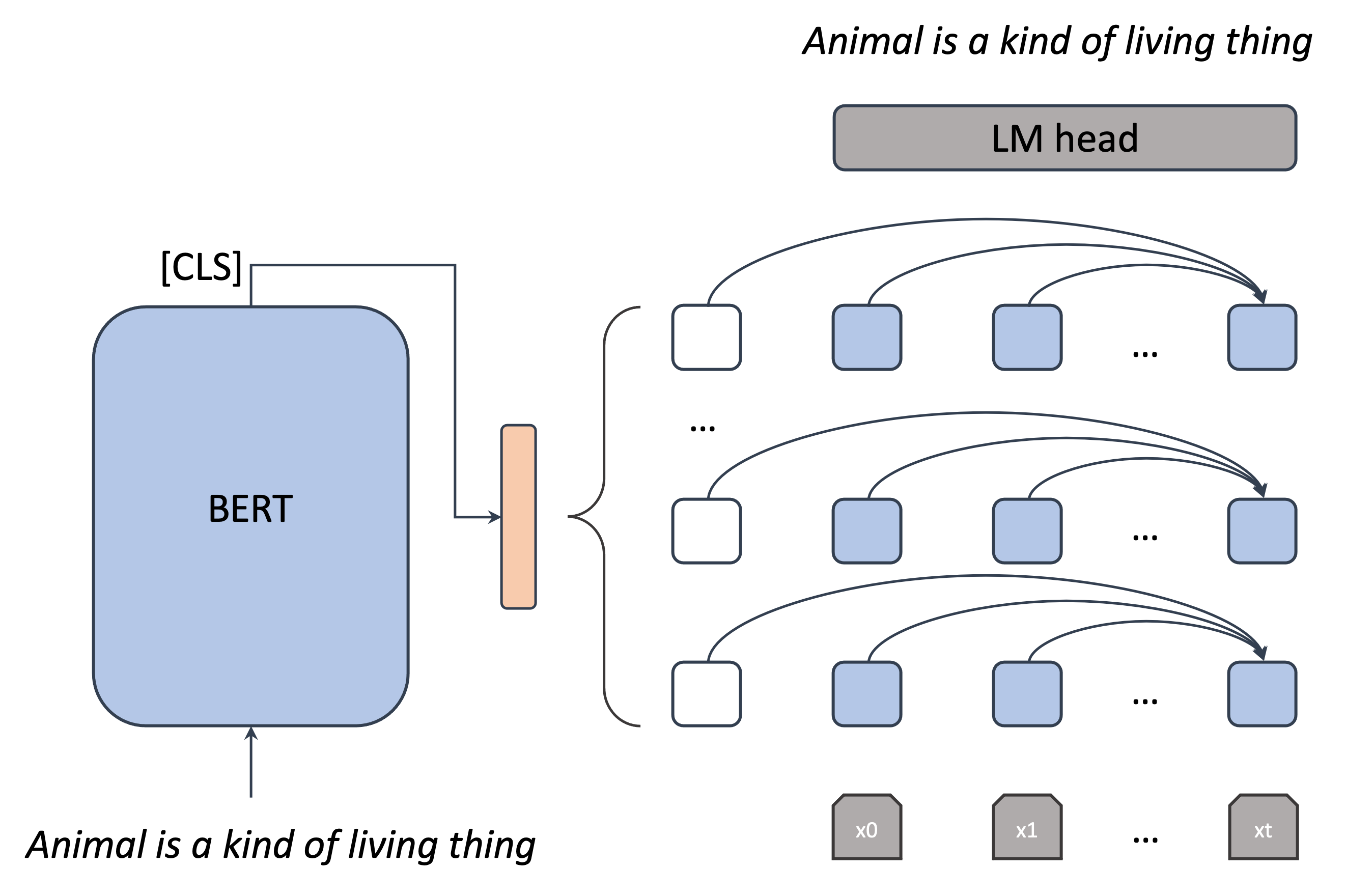}
    \caption{Latent sentence injection.}
    \label{fig:optimus}
 \end{center}
\end{figure}

\begin{table*}[ht!]
    \small
    \centering
\renewcommand\arraystretch{1.1}
    \begin{tabular}{|p{2cm}p{2cm}p{11cm}|}  \toprule
        \textbf{Semantic Tags}     & \textbf{Prop. \%}    & \textbf{Description and Example}                       \\ \hline
        ARGM-DIR & 0.80 & Directionals. E.g. all waves transmit energy \textbf{from one place to another}  \\ \hline
        ARGM-PNC & 0.08 & Purpose. E.g. many animals blend in with their environment \textbf{to not be seen by predators} \\ \hline
        ARGM-CAU & 0.05 & Cause. E.g. cold environments sometimes are white in color \textbf{from being covered in snow} \\ \hline
        ARGM-PRP & 1.30 & Purpose. E.g. a pot is made of metal \textbf{for cooking} \\ \hline
        ARGM-EXT & 0.04 & Extent. E.g. as the amount of oxygen exposed to a fire increases the fire will burn \textbf{longer} \\ \hline
        ARGM-LOC & 4.50 & Location. E.g. a solute can be dissolved \textbf{in a solvent} when they are combined \\ \hline
        ARGM-MNR & 2.00 & Manner. E.g. fast means \textbf{quickly} \\ \hline
        ARGM-MOD & 9.80 & Modal verbs. E.g. atom \textbf{can} not be divided into smaller substances \\ \hline
        ARGM-DIS & 0.07 & Discourse. E.g. if something required by an organism is depleted \textbf{then} that organism must replenish that something \\ \hline
        ARGM-GOL & 0.20 & Goal. E.g. We flew \textbf{to Chicago} \\ \hline
        ARGM-NEG & 1.20 & Negation. E.g. cactus wrens building nests in cholla cacti does \textbf{not} harm the cholla cacti \\ \hline
        ARGM-ADV & 6.70 & Adverbials \\ \hline
        ARGM-PRD & 0.20 & Markers of secondary predication. E.g. \\ \hline
        ARGM-TMP & 7.00 & Temporals. E.g. a predator \textbf{usually} kills its prey to eat it \\ \hline
        O & - & Empty tag. \\ \hline
        V & 100 & Verb. \\ \hline
        ARG0 & 32.0 & Agent or Causer. E.g. \textbf{rabbits} eat plants \\ \hline
        ARG1 & 98.5 & Patient or Theme. E.g. rabbits eat \textbf{plants} \\ \hline
        ARG2 & 60.9 & indirect object / beneficiary / instrument / attribute / end state. E.g. animals are \textbf{organisms} \\ \hline
        ARG3 & 0.60 & start point / beneficiary / instrument / attribute. E.g. sleeping bags are designed \textbf{to keep people warm} \\ \hline
        ARG4 & 0.10 & end point. E.g. when water falls from the sky that water usually returns \textbf{to the soil} \\ \toprule
        
    \end{tabular}
    \caption{Semantic Role Labels that appears in explanations corpus.} 
    \label{tab:srl_silva}
\end{table*}
\paragraph{Architecture.} Figure \ref{fig:optimus} provides a visual representation of the connection between BERT and GPT2 within the AutoEncoder architecture.

To train the CVAE, we use a new embedding layer for semantic roles and separate MLP layers $W^{srl}_{\mu}$ and $W^{srl}_{\sigma}$ to learn prior distribution.

\paragraph{Hyperparameters.} The training process of the decision tree binary classifier can be implemented via scikit-learn packages with default hyperparameters. As for Optimus, the latent space size is 32 in the experiment. The training details are following the original experiment from Optimus \cite{li2020optimus}.

\section{Further Experimental Results} \label{sec:apd_exp_res}

\begin{figure}[ht!]
\begin{center}
    \includegraphics[width=\columnwidth]{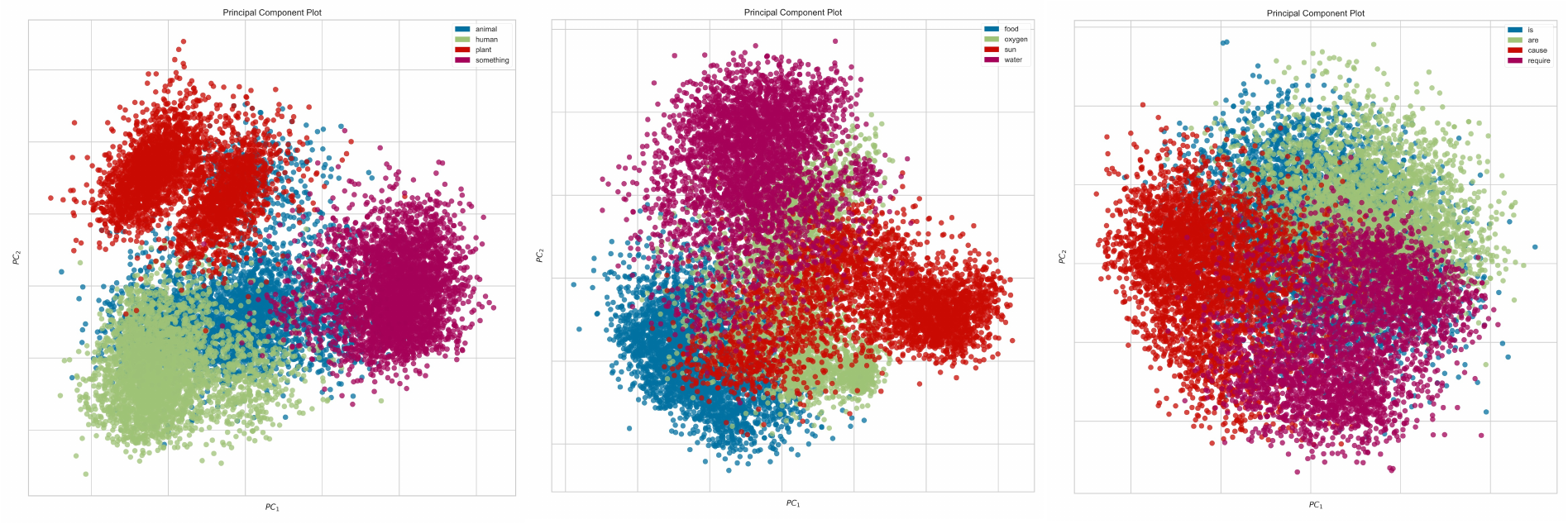}
    \caption{PCA visualisation.}
    \label{fig:pca}
 \end{center}
\end{figure}

\paragraph{Traversal visualisation.} PCA plots for ARG0, ARG1, and PRED are provided in Figure \ref{fig:pca}.

\begin{figure}[ht!]
\begin{center}
    \includegraphics[width=\columnwidth]{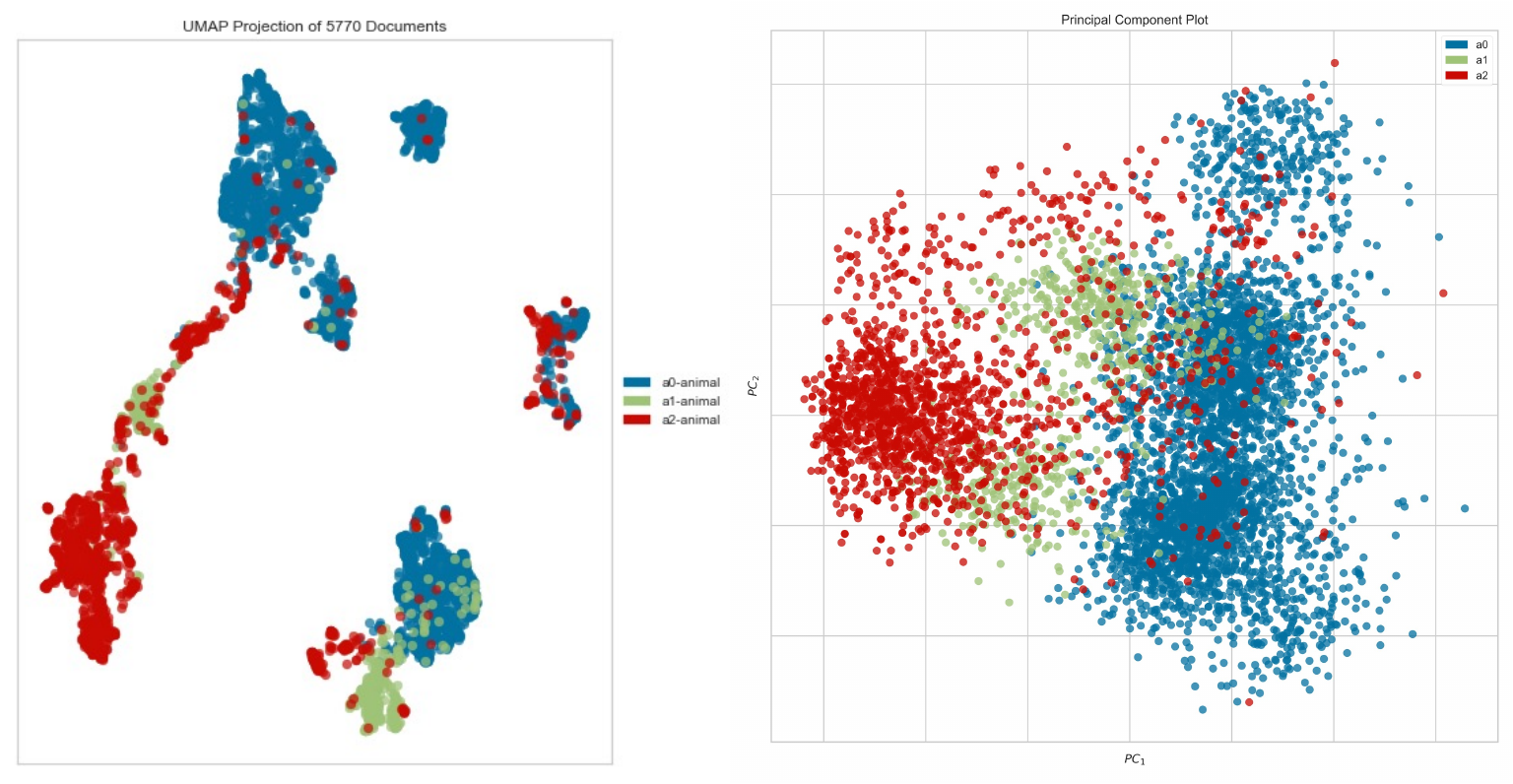}
    \caption{Visualisation for \textit{animal-ARG0,1,2}.}
    \label{fig:pca1}
 \end{center}
\end{figure}

In addition, we also provide the visualisation of word content \textit{animal} with different semantic roles: ARG0, ARG1, ARG2, in Figure \ref{fig:pca1}. From it, we can observe that the same content with different semantic roles can also be clustered and separated in latent space.


\paragraph{Qualitative evaluation for arithmetic.} Table~\ref{tab:arith_other_examples} lists the traversed explanations after addition (blue) and subtraction (red) on different semantic role information. We can observe that the resulting sentences after addition can hold the same role-content as inputs, revealing latent space geometry.

\begin{table*}[ht!]
\begin{tcolorbox}[fontupper=\small, fontlower=\small, middle=0.3cm, title=ADD and SUB arithmetic]
ARGUMENT1: \\
\underline{a needle is a kind of object} \\
\underline{a tire is a kind of object} \\
\\
a wire \textcolor{blue}{is a kind of object} \\
a stick \textcolor{blue}{is a kind of object} \\
a ball \textcolor{blue}{is a kind of object} \\
 \\
a serotype is \textcolor{red}{similar to intersex egg} \\
a zygote contains \textcolor{red}{many cell types} \\
an xylem is made \textcolor{red}{of two clumps} \\
\\
VERB: \\
\underline{chromosomes are located in the cells} \\
\underline{Australia is located in the southern hemisphere} \\
\\
stars are \textcolor{blue}{located} in the solar system \\
Jupiter is \textcolor{blue}{located} in the milky way galaxy \\
aurora is \textcolor{blue}{located} in the constellation of Leo \\
\\
a crystal is \textcolor{red}{made} of metal \\
an alloy is \textcolor{red}{made} of iron and zinc \\
an aluminum plug \textcolor{red}{is} nonmagnetic \\
\\
LOCATION: \\
\underline{volcanoes are often found under oceans} \\
\underline{mosquitos can sense carbon dioxide in the air} \\
\\
polar ice sheets are located \textcolor{blue}{along rivers} \\
hurricanes occur frequently along the coast \textcolor{blue}{in Africa} \\
tide waves cause flooding \textcolor{blue}{in coastal waters} \\
\\
\textcolor{red}{valley is a kind of location} \\
\textcolor{red}{shape is a property of rocks} \\
\textcolor{red}{desert is a kind of place} \\
\\
TEMPORAL: \\
\underline{as the population of prey decreases competition} \underline{between predators will increase} \\
\underline{as competition for resources decreases the ability} \underline{to compete for resources will increase} \\
\\
\textcolor{blue}{as the population of an environment decreases} ecosystem function will decrease \\
\textcolor{blue}{as the spread of available air mass increases} the population will increase \\
\textcolor{blue}{as the number of heavy traffic required increases} the traffic cycle will decrease \\
\\
\textcolor{red}{some types of lizards live in water} \\
\textcolor{red}{a rose is rich in potassium} \\
\textcolor{red}{a fern grass roots foot trait means a fern grass} \\
\\
NEGATION: \\
\underline{pluto has not cleared its orbit} \\
\underline{sound can not travel through a vacuum} \\
\\
radio waves \textcolor{blue}{don't} have electric charge \\
electromagnetic radiation \textcolor{blue}{does not} have a neutral electric charge \\
electromagnetic radiation contains \textcolor{blue}{no} electric charge \\
\\
Mars \textcolor{red}{is} a kind of moon / planet \\
Anothermic rock \textcolor{red}{is} a kind of metamorphic rock \\
Anal Cetus's skeleton \textcolor{red}{is} a kind of fossil
\end{tcolorbox}
\caption{Latent sapce arithmetic for five semantic tags (blue: addition, red: subtraction).}
\label{tab:arith_other_examples}
\end{table*}




\paragraph{Quantitative evaluation for arithmetic.} \label{sec:apd_consistency} Quantitative evaluation for our hypotheses via latent arithmetic. Both VERB and Object can perform high ratio after addition, indicating role-content separability.







\begin{figure*}[ht]
\begin{center}
    \includegraphics[width=\linewidth]{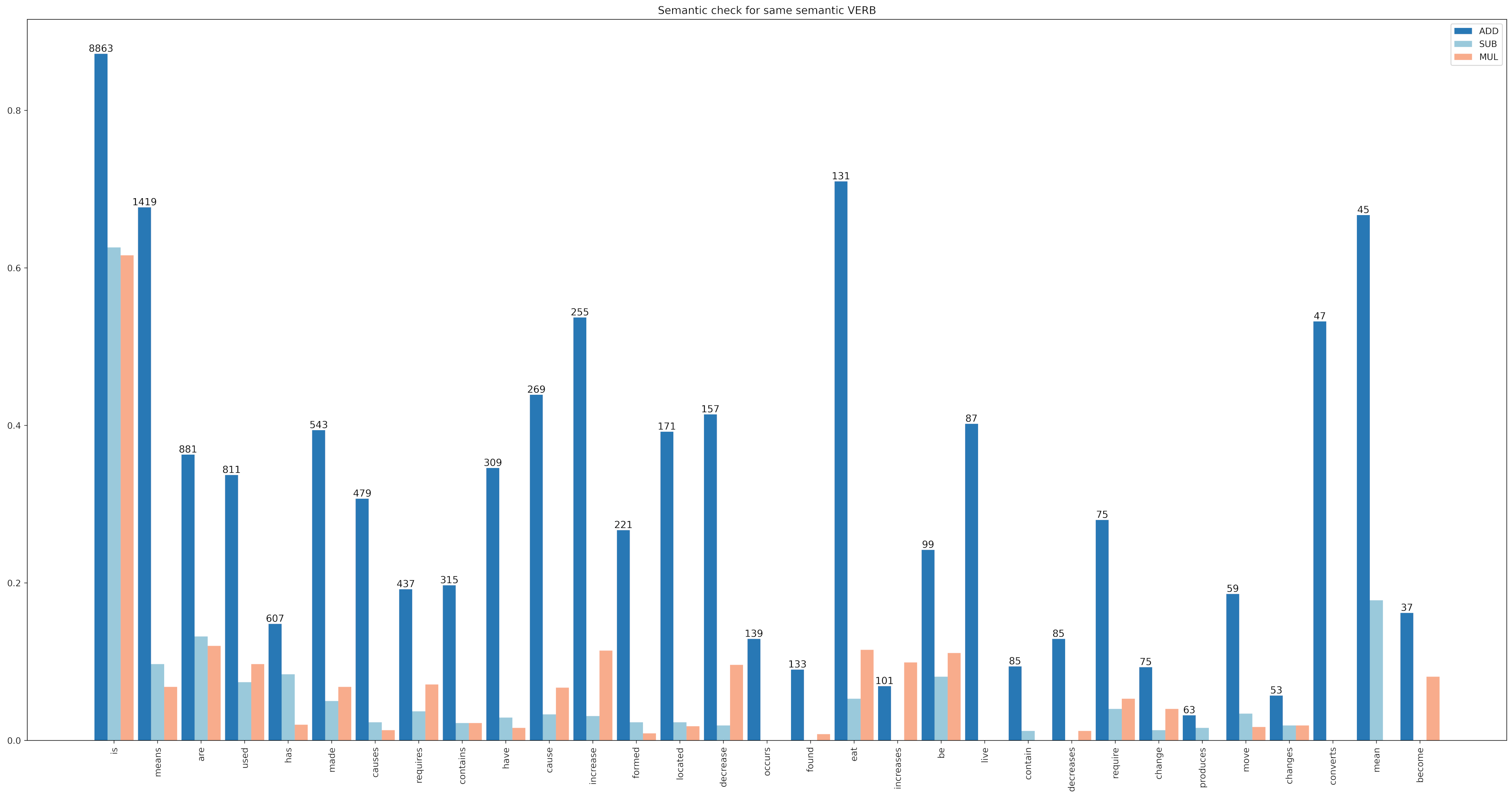}
    \caption{Predicate (VERB). The content \textit{is} shows the high ratio after subtraction, indicating that the \textit{V-is} is widely distributed over the latent space.}
    \label{fig:consistency_verb_sem}
    \end{center}
\end{figure*}

\begin{figure*}[ht]
\begin{center}
    \includegraphics[width=\linewidth]{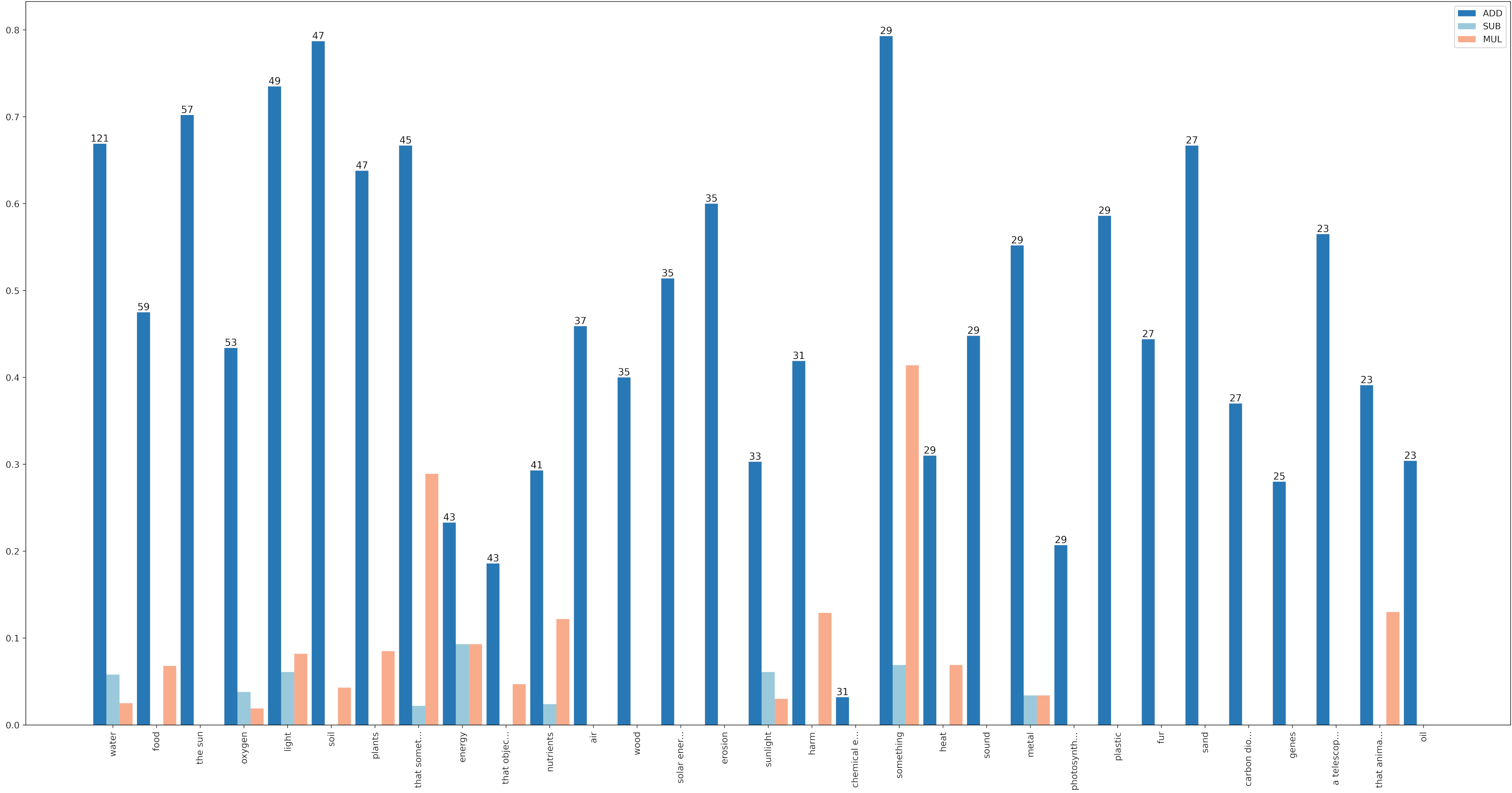}
    \caption{Object (ARG1).}
    \label{fig:consistency_arg1_sem}
\end{center}
\end{figure*}

\begin{figure*}[ht]
\begin{center}
    \includegraphics[width=\linewidth]{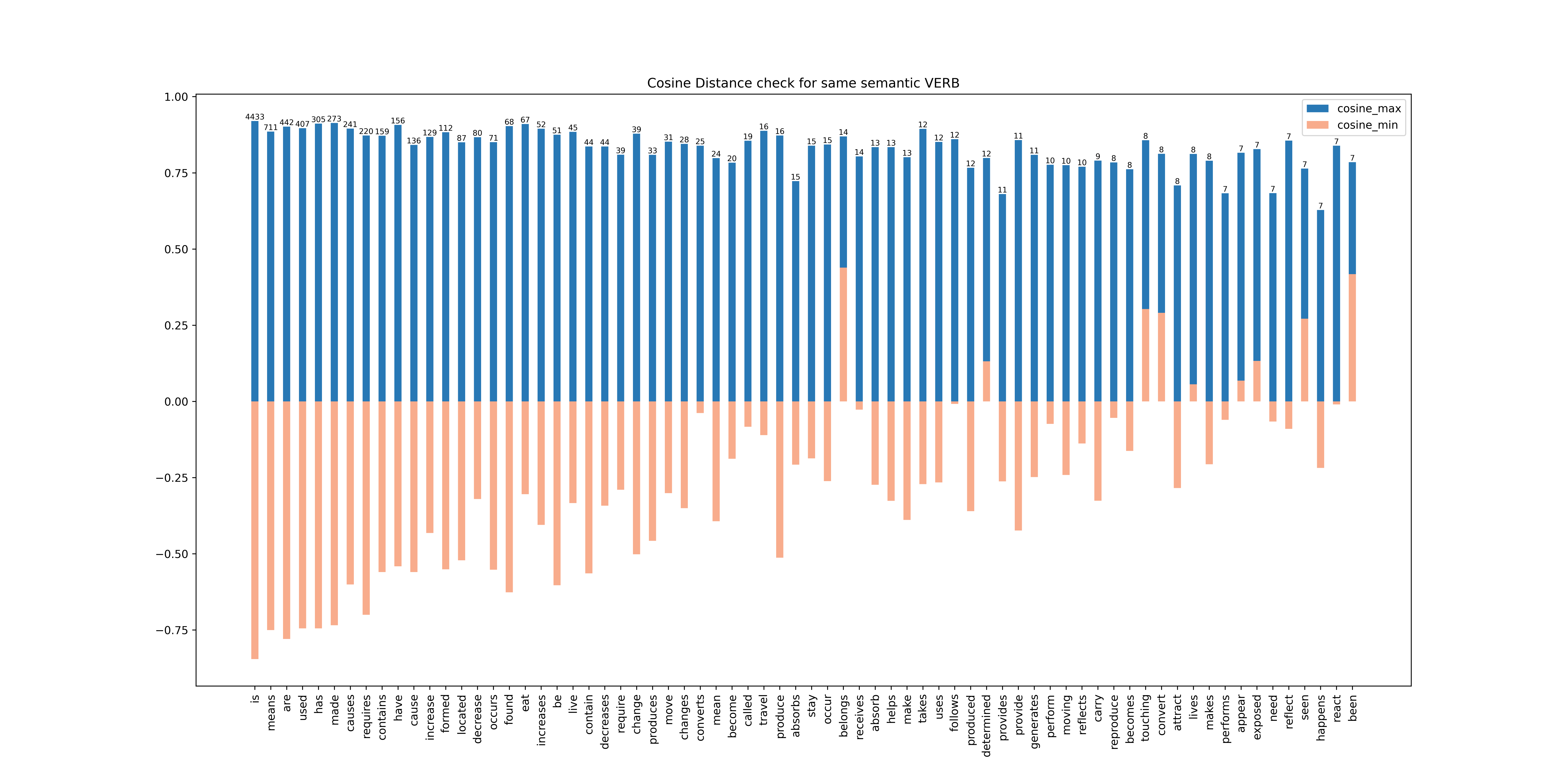}
    \caption{Cosine distance of sentence pairs in VERB-content clusters.}
    \label{fig:cos_dist_v}
    \end{center}
\end{figure*}

\begin{figure*}[ht]
\begin{center}
    \includegraphics[width=\linewidth]{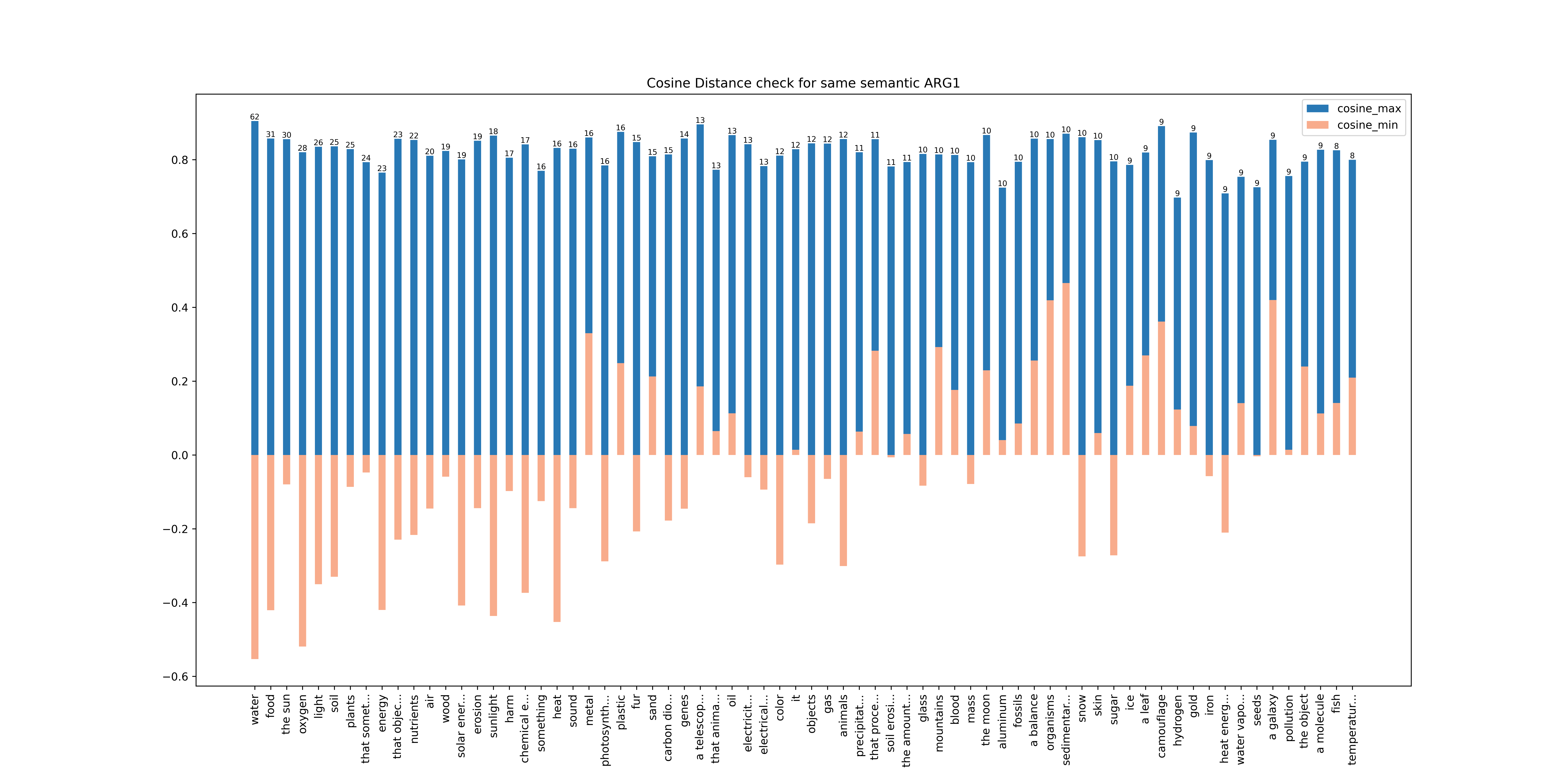}
    \caption{Cosine distance of sentence pairs in ARG1-content clusters.}
    \label{fig:cos_dist_a1}
\end{center}
\end{figure*}




\end{document}